\documentclass[letterpaper, 10 pt, conference]{ieeeconf}  % Comment this line out if you need a4paper

\IEEEoverridecommandlockouts                              % This command is only needed if 
\usepackage{graphicx}
\usepackage{multirow}
\usepackage{setspace}
\usepackage{gensymb}
\usepackage{cite}
\usepackage[table,xcdraw]{xcolor}
\usepackage{booktabs}
\usepackage{circledtext}
\usepackage{hyperref}
\usepackage{microtype}
\usepackage{epsfig}
\usepackage{graphicx}
\usepackage{amsmath}
\usepackage{amssymb}
\usepackage{colortbl}
\usepackage{xcolor} 
\usepackage{fontawesome}
\usepackage{microtype}
\title{\LARGE \bf
CD-FKD: Cross-Domain Feature Knowledge Distillation for Robust
Single-Domain Generalization in Object Detection
}

\author{Junseok Lee$^{1,*}$, Sungho Shin$^{2,*}$, Seongju Lee$^{3}$, and Kyoobin Lee$^{3, \dagger}$% <-this % stops a space
\thanks{}% <-this % stops a space
\thanks{$^{*}$ These authors contributed equally to this work.}
\thanks{$^{1}$ Junseok Lee is with the Advanced Robotics Lab, LG Electronics, Seoul, Republic of Korea.}
\thanks{$^{2}$ Sungho Shin is with the Place AI Flatform, Naver, Gyeonggi-do, Republic of Korea}
\thanks{$^{3}$ Seongju Lee and Kyoobin Lee are with the Department of AI Convergence, Gwangju Institute of Science and Technology (GIST), Gwangju, Republic of Korea.}
\thanks{$^{\dagger}$ Corresponding author: Kyoobin Lee (E-mail: kyoobinlee@gist.ac.kr)}
}

\renewcommand{\thefigure}{\Roman{figure}}
\renewcommand{\thetable}{\Roman{table}}
\begin{document}

\maketitle
\thispagestyle{empty}
\pagestyle{empty}

%%%%%%%%%%%%%%%%%%%%%%%%%%%%%%%%%%%%%%%%%%%%%%%%%%%%%%%%%%%%%%%%%%%%%%%%%%%%%%%%
\begin{abstract}

Single-domain generalization is essential for object detection, particularly when training models on a single source domain and evaluating them on unseen target domains. Domain shifts, such as changes in weather, lighting, or scene conditions, pose significant challenges to the generalization ability of existing models. To address this, we propose Cross-Domain Feature Knowledge Distillation (CD-FKD), which enhances the generalization capability of the student network by leveraging both global and instance-wise feature distillation. The proposed method uses diversified data through downscaling and corruption to train the student network, whereas the teacher network receives the original source domain data. The student network mimics the features of the teacher through both global and instance-wise distillation, enabling it to extract object-centric features effectively, even for objects that are difficult to detect owing to corruption. Extensive experiments on challenging scenes demonstrate that CD-FKD outperforms state-of-the-art methods in both target domain generalization and source domain performance, validating its effectiveness in improving object detection robustness to domain shifts. This approach is valuable in real-world applications, like autonomous driving and surveillance, where robust object detection in diverse environments is crucial.

\end{abstract}

%%%%%%%%%%%%%%%%%%%%%%%%%%%%%%%%%%%%%%%%%%%%%%%%%%%%%%%%%%%%%%%%%%%%%%%%%%%%%%%%
\section{INTRODUCTION}
\label{sec:intro}
In recent years, deep learning-based object detection technologies have made remarkably advanced, playing a critical role in various visual perception tasks \cite{cai2021yolov4, park2022deep, lee2025automated, lee2025robust}. These technologies achieve strong performance when the training and testing data share the same domain distribution, but this is rarely the case in real-world scenarios. In fields such as autonomous driving, industrial automation, and video surveillance, environmental changes caused by lighting, weather, or time of day result in domain shifts, which significantly degrade the performance of trained models \cite{geirhos2018generalisation, recht2019imagenet}.

To address this, methodologies such as unsupervised domain adaptation (UDA) \cite{li2022cross, chen2020harmonizing} and domain generalization (DG) \cite{carlucci2019domain, huang2020self} have been proposed. UDA aligns data distributions between the source and target domains to ensure model performance. However, it requires access to target domain data, limiting its ability to generalize across environments. DG focuses on training models to maintain performance on unseen target domains by leveraging diverse source domain data, making it a more challenging problem compared with UDA. Traditional DG methods aim to learn shared representations across multiple source domains or enhance data diversity via augmentation. However, these approaches often need multiple source domains, which can be impractical owing to time and cost constraints.

\begin{figure}[t!]
\centering
  \includegraphics[width=0.91 \columnwidth]{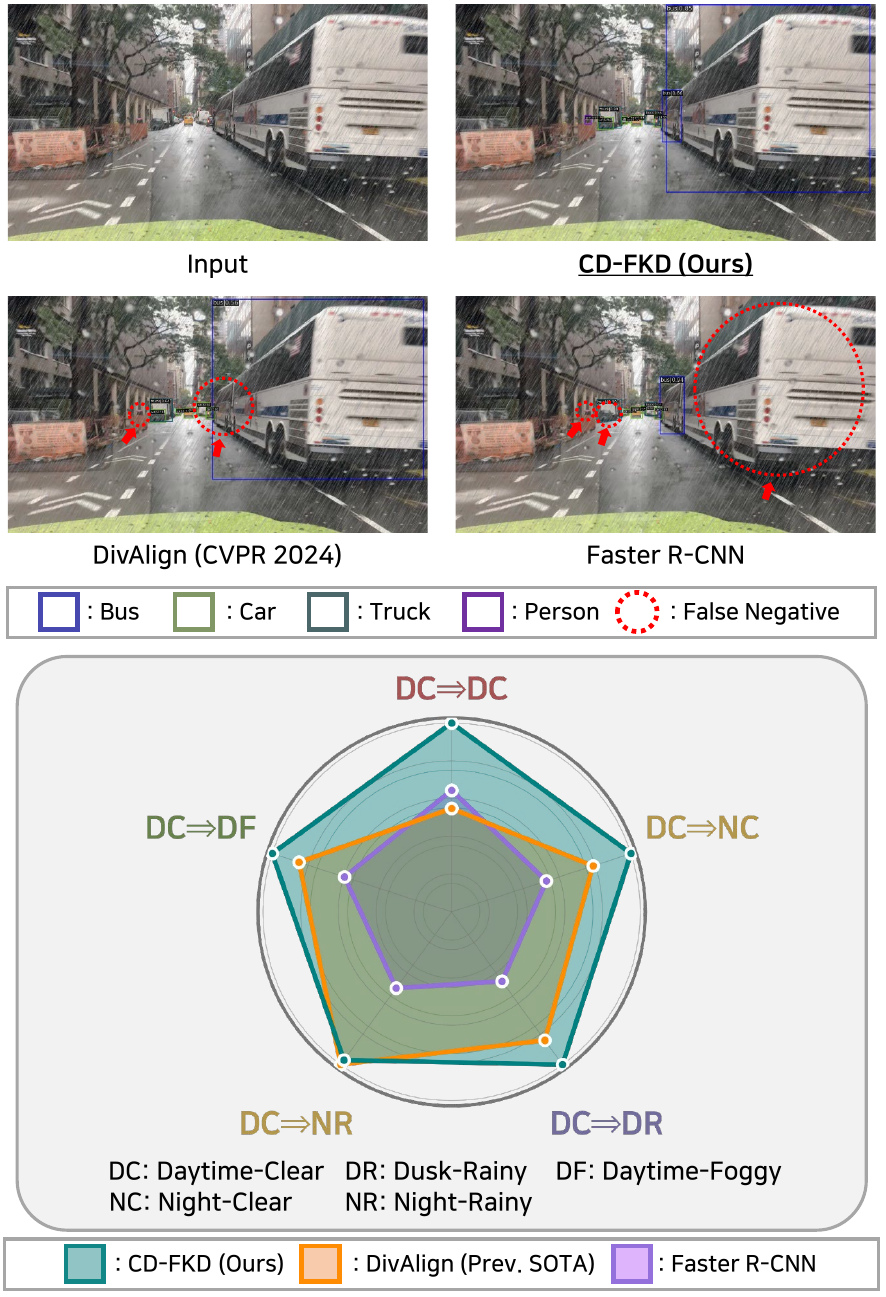}
%   \caption{Overview of results of our proposed CD-FKD. The left panel qualitatively compares our
% method with DivAlign \cite{danish2024improving} and Faster R-CNN \cite{ren2016faster}, on an Dusk-Rainy. The right panel shows a radar chart comparing relative
% performance of CD-FKD with DivAlign \cite{danish2024improving} (previous state-of-the-art) and Faster R-CNN \cite{ren2016faster}.} 
\caption{Overview of results of our proposed CD-FKD. The top panel qualitatively compares our method with DivAlign \cite{danish2024improving} and Faster R-CNN \cite{ren2016faster}, on an Dusk-Rainy. The bottom panel shows a radar chart comparing relative performance.} 
  \label{fig:figure1}
\end{figure}

Single-domain generalization (SDG) focuses on improving performance across diverse target domains using only a single source domain, offering a practical solution to the challenges of domain generalization \cite{wang2021learning, wan2022meta}. Unlike domain generalization (DG) methods, which require multiple source domains or domain-level annotations and thus increase complexity, SDG reduces both data requirements and costs. 

Traditional approaches typically rely on data augmentation and feature disentanglement. Feature disentanglement separates domain-invariant (object-centric) features from domain-specific (background-centric) features, using only the invariant features for object detection. However, this neglects the surrounding background, limiting the detector’s ability to understand the full context of the image. Although data augmentation can improve generalization across environments, it has been shown to reduce performance on the source domain \cite{kirichenko2024understanding}. To address these issues, the proposed method enables the detector to extract object-centric features while maintaining a comprehensive understanding of the image’s context. This not only enhances generalization performance on target domains but also improves detection on the source domain.

We propose a novel methodology that effectively operates across both the source and target domains by understanding causal features and learning the global context of images. To achieve this, we introduce \textbf{C}ross-\textbf{D}omain \textbf{F}eature \textbf{K}nowledge \textbf{D}istillation (CD-FKD) for single domain generalized object detection. The proposed method comprises two core components: (1) cross-domain knowledge distillation (KD) framework and (2) cross-domain feature distillation loss. The cross-domain KD framework bridges domain gaps between teacher and student networks. The teacher network receives original source domain data, whereas the student network learns from diversified source data with various scales and corruptions.

The primary goal of CD-FKD is to train the network to effectively extract object-centric features even in data with various distortions. The proposed feature distillation process comprises two key steps: (1) global feature distillation, which enables the network to learn the global context of an image and (2) instance-wise feature distillation, which helps the network focus on object-specific features. The instance-wise feature distillation component ensures that the network effectively extracts object-centric features, whereas the global feature distillation component focuses on learning the overall context of the image. These distillation techniques prevent the network from overfitting to a specific domain and enable robust learning under various distortion conditions. Consequently, the student network is capable of performing reliably under diverse distortions while maintaining strong performance on the source domain data. This approach enhances the generalization between the source and target domains, enabling robust object detection across a wide range of domain environments. As illustrated in Figure \ref{fig:figure1}, the proposed method achieves superior SDG performance compared with state-of-the-art (SOTA) methods. In summary, the main contributions of this study are as follows: 
\begin{itemize}
\item We propose CD-FKD, a novel cross-domain FKD method for single-domain generalized object detection.
\item We evaluate the proposed method against other SDG approaches on the SDG benchmark dataset and demonstrate that the proposed method outperforms previous methods. Also, we present comprehensive analyses that contribute to an improved understanding of the proposed framework. 
% \item We present comprehensive analyses that contribute to an improved understanding of the proposed framework.
\end{itemize}

\section{Related Work}
\subsection{Single Domain Generalization}
SDG focuses on training models with a single source domain while ensuring they generalize well to unseen target domains, a key task when collecting data from multiple domains is difficult. SDG methods are divided into data augmentation \cite{volpi2018generalizing, wang2021learning} and domain-invariant feature learning \cite{wu2022single}. Data augmentation generates synthetic samples to improve robustness against distribution shifts, whereas domain-invariant feature learning aims to extract stable features across domains.

Recent studies on single-domain generalized object detection have progressed with methods such as S-DGOD \cite{wu2022single}, which uses contrastive learning and self-distillation to improve generalization. CLIP-Gap \cite{vidit2023clip} leverages CLIP for text-based augmentations to enhance object detection without requiring target domain data. G-NAS \cite{wu2024g} reduces overfitting using differentiable NAS and OoD-aware G-loss. UFR \cite{liu2024unbiased} improves generalization with scene-level causal attention and object-level prototypes. Danish et al. \cite{danish2024improving} introduced carefully selected augmentations to diversify the source domain and a classification and localization alignment method to enhance out-of-domain detections. Li et al. \cite{li2024prompt} proposed a dynamic object-centric perception network with prompt learning to adapt to image complexity and improve cross-domain generalization. Despite these advancements, we propose a novel cross-domain feature distillation approach within a knowledge distillation (KD) framework that outperforms existing methods.

\begin{figure*}[t!]
\centering
  \includegraphics[width=0.90 \textwidth]{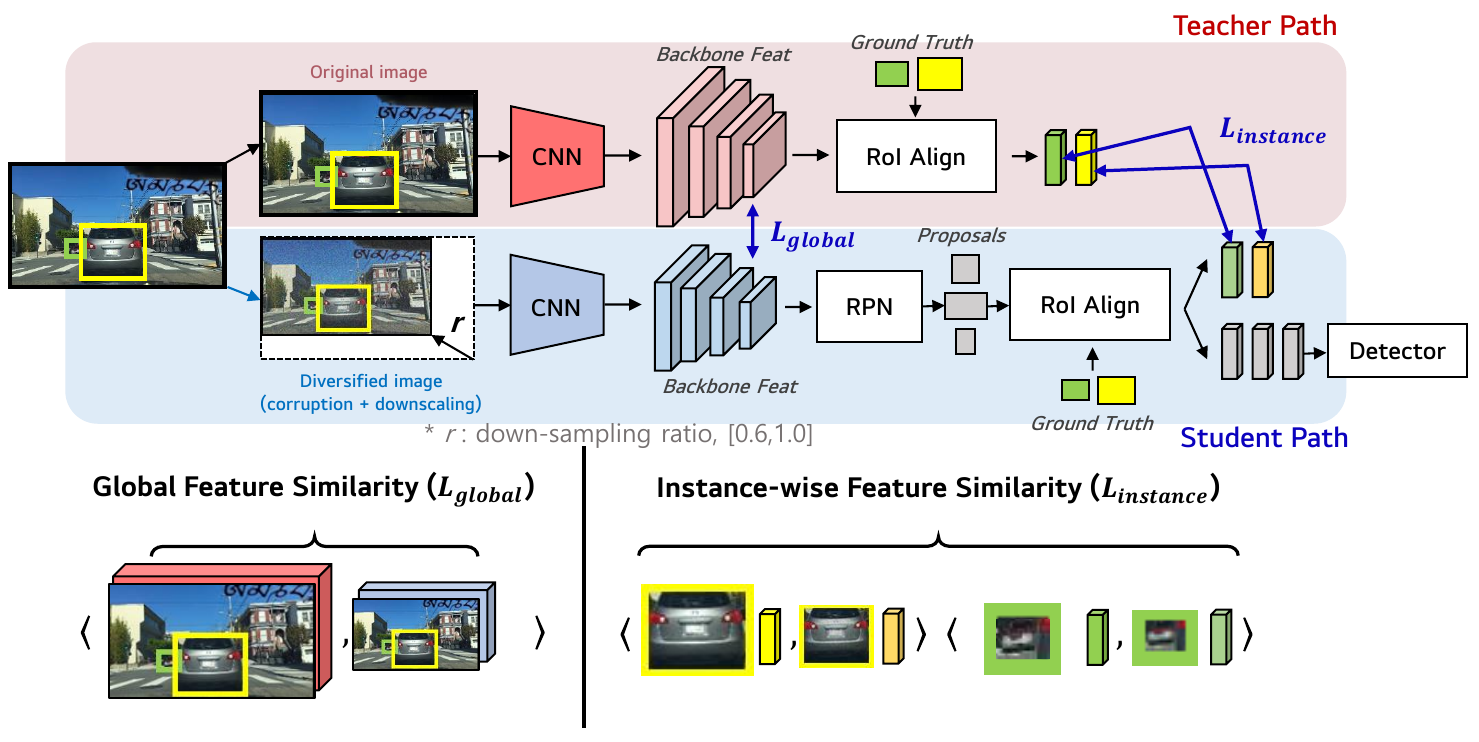}
  \caption{Illustration of the proposed single-domain generalized object detection using cross-domain FKD. As a KD framework, the frozen teacher network (represented in pink color) receives source domain data, whereas the student network (represented in blue sky color) is provided with downscaled and corrupted source domain data.} 
  \label{fig:CD-FKD}
\end{figure*}

\subsection{Knowledge Distillation for Object Detection}
KD was originally proposed for model compression in image classification, transferring knowledge from large teacher models to smaller student models \cite{hinton2015distilling}. Initially designed for image classification \cite{shin2022teaching, el2024hdkd}, KD has been expanded to object detection for both compression and performance enhancement \cite{ bang2024radardistill, zhao2024crkd}. By replicating the capabilities of the teacher model with fewer resources, KD has proven to be a versatile strategy.

In object detection, KD improves compact model performance by enabling them to emulate key features from larger models. FineGrained \cite{wang2019distilling} introduces feature imitation for performance enhancement, whereas DeFeat \cite{guo2021distilling} separates foreground and background features for independent distillation. FGD \cite{yang2022focal} and ScaleKD \cite{zhu2023scalekd} improve detection accuracy by leveraging focal-global features and scale-aware knowledge transfer, respectively. CrossKD \cite{wang2024crosskd} uses cross-head distillation to improve performance by mimicking teacher predictions.

We propose a KD approach for SDG in object detection. The proposed method allows the student model to detect objects in images with reduced visibility and noise by mimicking the ability of the teacher model to detect objects in clear images. This improves performance under challenging conditions, making the model robust for unseen target domains.

\section{Method}
We introduce CD-FKD, a simple yet effective single-domain generalized object detection framework. First, we provide an overview of the proposed framework in Section \ref{subsec:framework}. Second, we introduce the method for generating diversified source domain image in Section \ref{subsec:div}. Third, we describe the global feature distillation method in Section \ref{subsec:feature}. Finally, we present the instance-specific feature distillation approach in Section \ref{subsec:instance}.

\subsection{Cross Domain Distillation Framework}
\label{subsec:framework}
The proposed framework uses a self-distillation structure with two identical detectors. We employ the widely recognized Faster R-CNN, a popular two-stage detector for single-domain object detection. The key idea is to leverage the differences in input images between the teacher and student networks, maximizing the benefits of this approach. In this framework, the teacher network receives clear, high-resolution images, enabling it to extract detailed features. In contrast, the student network is trained with corrupted and downscaled images, which makes detecting small objects more challenging. Despite these difficulties, the student network learns to handle such scenarios, becoming more robust. Training on these challenging inputs exposes the student network to difficult detection situations, promoting more rigorous learning. Through distillation, the student learns feature representations from the teacher, improving its ability to detect fine-grained details in tough environments. This process helps the student network become more resilient to corruption and better at small object detection.

As shown in Figure \ref{fig:CD-FKD}, the teacher network receives original source domain data, while the student network is trained on diversified source domain data with various scales and corruptions. The training begins by pre-training the teacher network on the source domain data. During distillation, the teacher network’s parameters are frozen, and it performs inference on the clear source domain data. Meanwhile, the student network learns object classification and localization from the diversified data, benefiting from the teacher’s knowledge.

\begin{figure}
  \centering
  \includegraphics[width=0.90\columnwidth]{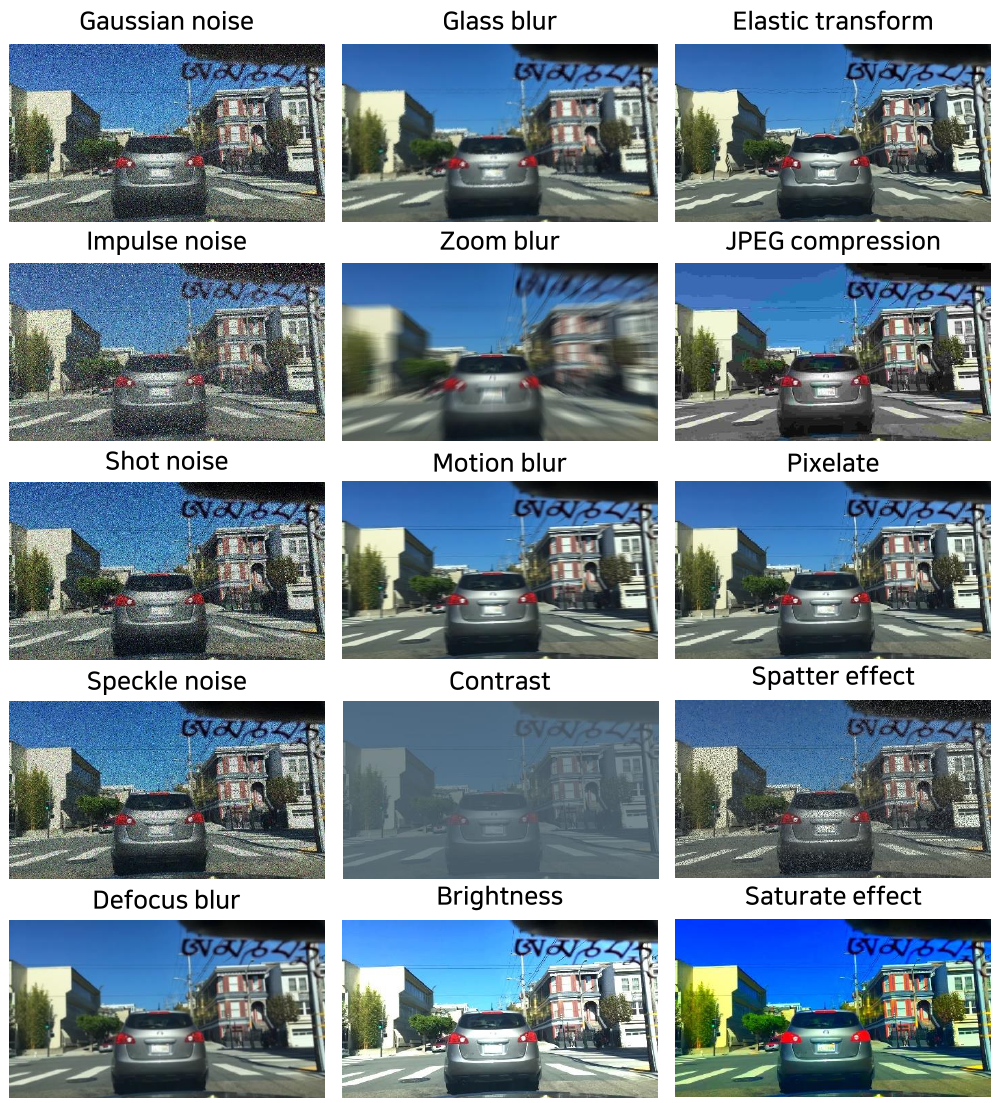}
  \caption{Examples of corrupted source domain data} 
  \label{fig:example_corrupt}
\end{figure}

\subsection{Diversified Source Domain Data}
\label{subsec:div}
The proposed approach leverages diversified source domain images ($\textit{D}_{\varphi}$) by applying various downscaling and corruption techniques to the source domain images ($\textit{D}_{s}$). Object detectors face challenges in target domains due to significant differences from the source domain, such as changes in image quality and object sizes. To address this, the proposed method applies downscaling and corruption to create discrepancies between the source and target domains.

We downscale the source image resolution to generate low-resolution images, while preserving the original high-resolution image for the teacher network. The teacher network receives the high-resolution image, and the student network is trained with the downsampled images. This encourages the student network to learn to detect small objects effectively, even from low-resolution data. Additionally, various corruptions are applied to the source domain images during training, enhancing generalization and preventing overfitting, as demonstrated in previous works like SDG \cite{danish2024improving}. These corruptions, such as object occlusion and image blurring, can challenge the network’s ability to detect objects. To mitigate this, the teacher network receives clear source domain data, while the student network is trained on corrupted data. This setup allows the student network to learn from noisy data, extracting features that reflect those learned by the teacher network, ultimately improving generalization to unseen target domains.

We apply generic corruptions that are independent of the target domains. As shown in Figure \ref{fig:example_corrupt}, a total of 15 different corruptions are applied using ImageNet-C \cite{hendrycks2019robustness}. These corruptions are applied with equal probability, and the intensity levels (ranging from 1 to 5) are distributed evenly across the augmentations.

\subsection{Global Feature Distillation}
\label{subsec:feature}
Global feature distillation is a technique that guides the student network to focus on important areas related to object detection, rather than focusing on the noise in diversified images. Diversified images distort object features, leading to semantic mismatches in the feature space. To address this, the student network is trained to learn semantically consistent features from diversified images.

During the distillation process, the features extracted by the teacher network from the original image (${x}^{s}$) serve as a reference for training the student network. The student network is trained to align its features, extracted from the diversified image (${x}^{\varphi}$), with this reference. Specifically, we obtain the backbone features of the teacher, denoted as $F^{s}_{T}$ = $f_\text{T}({x}^{s})$, and the backbone features of the student, denoted as $F^{\varphi}_{S}$ = $f_\text{S}({x}^{\varphi})$, where $f_\text{T}(\cdot)$ and $f_\text{S}(\cdot)$ represent the backbones of the teacher and student networks, respectively. To align $F^{\varphi}_{S}$ with $F^{s}_{T}$, a global feature distillation loss function ($\mathcal{L}_{\textit{global}}$) is employed.

In the global feature distillation process, feature embeddings ($F^{s}_{T}$, $F^{\varphi}_{S}$) are extracted from the final backbone layers of both the teacher and student networks, which use ResNet-101. $F^{s}_{T}$ has a fixed size, whereas $F^{\varphi}_{S}$ is resized according to the scale-down ratio (0.6 to 1.0) of the input resolution. To align the feature map sizes of $F^{\varphi}_{S}$ and $F^{s}_{T}$, bilinear interpolation is applied to $F^{\varphi}_{S}$ to match its size with $F^{s}_{T}$.

The two feature embeddings are flattened, and cosine similarity loss is applied to maximize the cosine similarity between the two features. The cosine similarity between $F^{T}_{s}$ and $F^{S}_{\varphi}$ is calculated using the formula below:

\begin{equation}
\mathcal{L}_{\text{\textit{global}}}(F^{s}_{\text{T}}, F^{\varphi}_{\text{S}}) = \sum_{i=1}^{N} \left( 1 - \frac{F^{s}_{\text{T},i} \cdot F^{\varphi}_{\text{S},i}}{\|F^{s}_{\text{T},i}\| \cdot \|F^{\varphi}_{\text{S},i}\|} \right)
\end{equation}
where \textit{N} denotes the number of train images, \textit{i}-th refers to the image index.

\begin{table*}[!h]
\centering
\footnotesize
\caption{Single domain generalization object detection results. Average results are calculated using the results from four target domains to compare the generalization ability. The results of the compared models are obtained from their respective papers. \textbf{Bold}/\underline{underline} indicates the best/second-best result. 
} \label{tab:main_result}
\begin{tabular}{l|ccccc|c}
\hline
Method        & Daytime-Clear & Night-Clear   & Dusk-Rainy    & Night-Rainy   & Daytime-Foggy & Average       \\ \hline
Faster R-CNN \cite{ren2016faster} & 54.9          & 36.6          & 27.9          & 12.1          & 32.1          & 27.2          \\
IBN-Net \cite{pan2018two}      & 49.7          & 32.1          & 26.1          & 14.3          & 29.6          & 25.5          \\
SW  \cite{pan2019switchable}          & 50.6          & 33.4          & 26.3          & 13.7          & 30.8          & 26.1          \\
IterNorm \cite{huang2019iterative}     & 43.9          & 29.6          & 22.8          & 12.6          & 28.4          & 23.4          \\
ISW \cite{choi2021robustnet}          & 51.3          & 33.2          & 25.9          & 14.1          & 31.8          & 26.3          \\
S-DGOD \cite{wu2022single}        & 56.1          & 36.6          & 28.2          & 16.6          & 33.5          & 28.7          \\
CLIP-Gap \cite{vidit2023clip}     & 51.3          & 36.9          & 32.3          & 18.7          & 38.5          & 31.6          \\
G-NAS \cite{wu2024g}        & 58.4          & \underline{45.0}    & 35.1          & 17.4          & 36.4          & 33.5          \\
PDDOC  \cite{li2024prompt}       & 53.6          & 38.5          & 33.7          & 19.2          & 39.1          & 32.6          \\
DivAlign \cite{danish2024improving}     & 52.8          & 42.5          & \underline{38.1}    & \textbf{24.1} & 37.2          & \underline{35.5}    \\
UFR   \cite{liu2024unbiased}        & \underline{58.6}    & 40.8          & 33.2          & 19.2          & \underline{39.6}    & 33.2          \\ \hline
\rowcolor[HTML]{d1d1d1} 
CD-FKD (Ours) & \textbf{62.7} & \textbf{47.3} & \textbf{42.3} & \underline{23.4}    & \textbf{40.2} & \textbf{38.3} \\ \hline
\end{tabular}%
\end{table*}

\subsection{Instance-Wise Feature Distillation}
\label{subsec:instance}
Instance-wise feature distillation is a technique that focuses on the relationship between objects in diversified images and their corresponding objects in original images, excluding the background. The student network is provided with an image that has been diversified compared with the original image. These diversified images often suffer from occlusions or blurring, and owing to the smaller object sizes, they lack the visual information needed for the detector to accurately recognize the objects. Therefore, the goal of instance-wise feature distillation is to address the problem of reduced object visibility in corrupted images. The main objective is to use the region of interest (RoI) to make the RoI features of the diversified image as similar as possible to those of the original image. To achieve this, instance-wise feature distillation guides the student network to learn to extract features from the diversified image that are similar to those in the original image.

To implement this approach, the backbone features of the teacher ($F^{s}_{T}$) and student ($F^{\varphi}_{S}$) are processed using RoI Align (RA) with the ground truth bounding box values (GT) as the RoIs. This process extracts the instance features ($I^{s}_{\text{T}}$) of the teacher and instance features ($I^{\varphi}_{\text{S}}$) of the student, which are defined as follows: $I^{s}_{\text{T}}$ = RA($F^{s}_{T}$, GT), $I^{\varphi}_{\text{S}}$ = RA($F^{\varphi}_{S}$, GT). $I^{\varphi}_{\text{S}}$ and $F^{\varphi}_{S}$ are then aligned to ensure that corresponding objects are properly matched. All aligned features are flattened, and a cosine similarity loss is applied to calculate the similarity between the corresponding objects. The cosine similarity between the object features in the teacher and student networks is calculated as follows:
\begin{equation}
\mathcal{L}_{\text{\textit{instance}}}(I^{s}_{\text{T}}, I^{\varphi}_{\text{S}}) = \sum_{i=1}^{N} \sum_{j=1}^{O} \left( 1 - \frac{I^{s}_{\text{T},i,j} \cdot I^{\varphi}_{\text{S},i,j}}{\|I^{s}_{\text{T},i,j}\| \cdot \|I^{\varphi}_{\text{S},i,j}\|} \right)
\end{equation}
where \textit{N} denotes the number of images, \textit{i}-th refers to the image index, \textit{O} is the number of instances in the image, and \textit{j}-th denotes to the instance index.

This approach enables the student network to extract features from diversified images guided by clean-image features, while global and instance-wise distillation ensures it captures both overall context and individual instance knowledge.

Finally, the proposed model comprises the sum of the loss function for object localization and classification ($\mathcal{L}_{\text{\textit{det}}}$), global feature distillation loss (${L}_{\text{\textit{global}}}$), and instance-wise feature distillation ($\mathcal{L}_{\text{\textit{instance}}}$).
Our overall training objective is as follows:
\begin{equation}
\mathcal{L}_{\text{\textit{total}}} = \mathcal{L}_{\text{\textit{det}}} + \alpha \mathcal{L}_{\text{\textit{global}}} + \beta \mathcal{L}_{\text{\textit{instance}}}
\end{equation}
where $\alpha$ and $\beta$ denote hyperparameters that contribute toward balancing $\mathcal{L}_{\text{\textit{global}}}$ and $\mathcal{L}_{\text{\textit{instance}}}$.

\begin{figure*}[t!]
\centering
  \includegraphics[width=1.0 \textwidth]{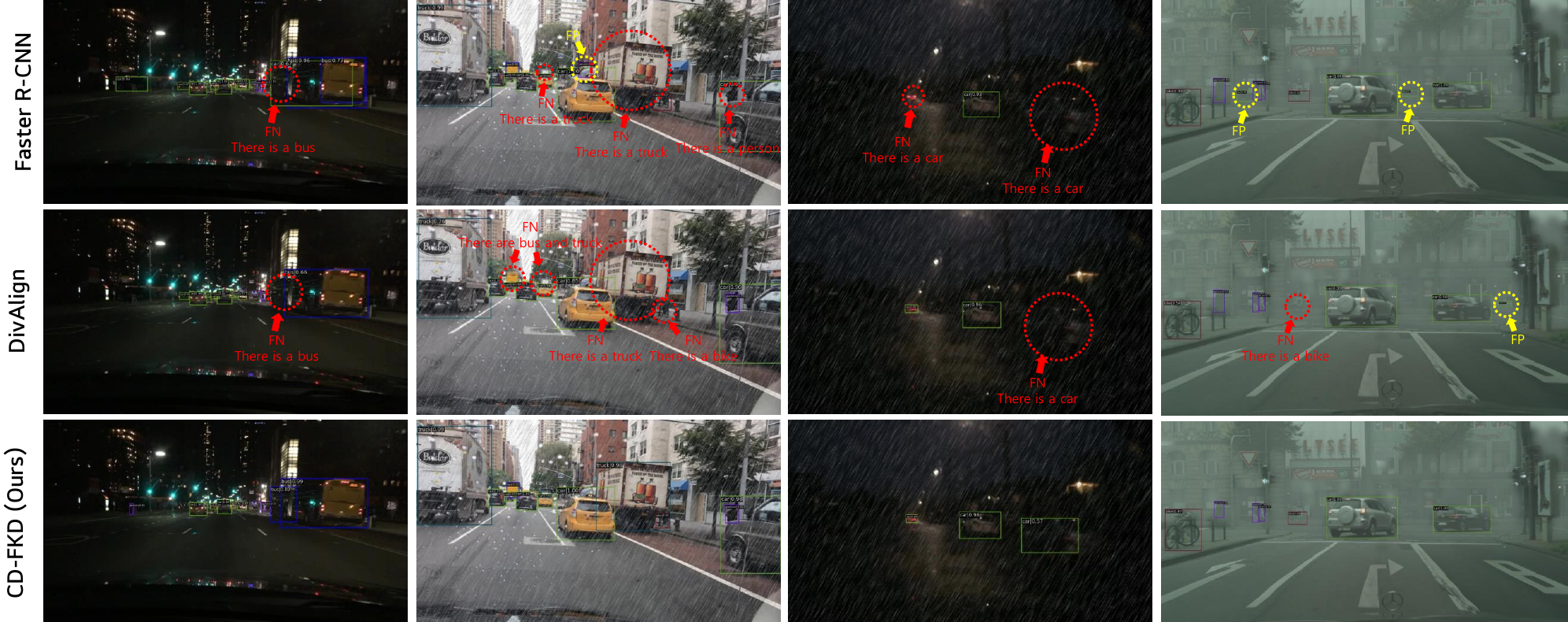}
  \caption{Qualitative evaluation results of the model's generalization ability on the Night-Clear, Dusk-Rainy, Night-Rainy, and Daytime-Foggy scenes. The top-row images show the results of Faster R-CNN \cite{ren2016faster}. The middle-row images show the results of DivAlign \cite{danish2024improving}. The bottom-row images show the results of our method. Red circles and arrows indicate false negatives, while yellow circles and arrows indicate false positives.} 
  \label{fig:vis}
\end{figure*}

\section{Experiments}

\subsection{Experimental Setup}
\textbf{Dataset.} 
We evaluated the proposed method on the diverse weather dataset, a single domain generalization benchmark dataset in urban scenes built by \cite{wu2022single}. The dataset comprises five different weather domains: Daytime-Clear, Night-Clear, Dusk-Rainy, Night-Rainy, and Daytime-Foggy. Daytime-Clear is used as the source domain for training, while the other four weather domains are used as target domains for testing. Daytime-Foggy comprises 19,395 images used for training and 8,313 images for testing. The four target domains used solely for evaluation are as follows: Night-Clear with 26,158 images, Dusk-Rainy with 3,501 images, Night-Rainy with 2,494 images, and Daytime-Foggy with 3,775 images. This dataset is an urban scene dataset containing seven object classes: bus, bike, car, motor, person, rider, and truck.

\textbf{Implementation Details.} We use Faster R-CNN \cite{ren2016faster} with a ResNet101-FPN \cite{lin2017feature} as the feature backbone, implemented in mmdetection \cite{mmdetection}, as the detector. The backbone is initialized with pre-trained weights from ImageNet. The model is trained using the stochastic gradient descent (SGD) optimizer with a learning rate of 0.01, momentum of 0.9, and weight decay of 0.0001. The batch size is set to 4. The values of $\alpha$ and $\beta$ in the loss function are both set to 1.0.

\subsection{Comparison with State-of-the-arts}
Our method is compared with normalization-based SDG works \cite{pan2018two, pan2019switchable, huang2019iterative, choi2021robustnet} and recent SOTA single-domain generalized detectors \cite{wu2022single, vidit2023clip, wu2024g, li2024prompt, danish2024improving, liu2024unbiased}. For the baseline, we use Faster R-CNN \cite{ren2016faster} initialized with ImageNet pre-trained weights. We evaluate the performance of the proposed method on the source domain, Daytime-Clear, and assess its generalization capability on unseen target domains, including Night-Clear, Dusk-Rainy, Night-Rainy, and Daytime-Foggy. The evaluation metric used is mean average precision (mAP), and we report results for mAP@0.5.

As summarized in Table \ref{tab:main_result}, the proposed method, CD-FKD, achieves the good performance not only on the source domain but also across all target domains. The average performance across the four target domains (Night-Clear, Dusk-Rainy, Night-Rainy, and Daytime-Foggy) achieves a mAP@0.5 of 38.3\%, representing an improvement of 11.1\% mAP@0.5 over the Faster R-CNN baseline. Furthermore, the proposed method outperforms the previous best, DivAlign \cite{danish2024improving}, by 2.8\% mAP@0.5. Remarkably, the proposed method enhances generalization to target domains without compromising performance on the source domain, surpassing previous methods on both source and target domains. These experimental results validate that the proposed approach significantly improves performance on the source domain while also enhancing generalization to unseen domains.

Figure \ref{fig:vis} visually compares the detection results of the Faster R-CNN \cite{ren2016faster}, DivAlign \cite{danish2024improving}, and the proposed method. The figure shows the detection results of each model for the Night-Clear, Dusk-Rainy, Night-Rainy, and Daytime-Foggy scenes from left to right. In contrast to the bottom-row results of the proposed method, the top-row Faster R-CNN and middle-row DivAlign results display both false negatives (indicated by red) and false positives (indicated by yellow).

\begin{table}[]
\centering
\setlength\tabcolsep{2.5pt}
\footnotesize
\caption{Quantitative results (\%) on the Night-Clear scene.
} \label{tab:night_clear}
\begin{tabular}{l|ccccccc|
>{\columncolor[HTML]{d1d1d1}}c }
\hline
Method       & bus           & bike          & car           & motor         & person        & rider         & truck         & mAP           \\ \hline
Faster R-CNN \cite{ren2016faster} & 37.5          & 36.6          & 62.0          & 14.0          & 43.6          & 28.8          & 41.6          & 37.7          \\
IBN-Net \cite{pan2018two}     & 37.8          & 27.3          & 49.6          & 15.1          & 39.2          & 27.1          & 38.9          & 32.1          \\
SW    \cite{pan2019switchable}       & 38.7          & 29.2          & 49.8          & 16.6          & 31.5          & 28.0          & 40.2          & 33.4          \\
IterNorm  \cite{huang2019iterative}   & 38.5          & 23.5          & 38.9          & 15.8          & 26.6          & 25.9          & 38.1          & 29.6          \\
ISW   \cite{choi2021robustnet}       & 38.5          & 28.5          & 49.6          & 15.4          & 31.9          & 27.5          & 41.3          & 33.2          \\
S-DGOD  \cite{wu2022single}     & 40.6          & 35.1          & 50.7          & 19.7          & 34.7          & 32.1          & 43.4          & 36.6          \\
CLIP-Gap  \cite{vidit2023clip}   & 37.7          & 34.3          & 58.0          & 19.2          & 37.6          & 28.5          & 42.9          & 36.9          \\
G-NAS   \cite{wu2024g}     & \underline{46.9}    & \textbf{40.5} & \underline{67.5}    & \underline{26.5}    & \underline{50.7}    & \underline{35.4}    & \underline{47.8}    & \underline{45.0}    \\
PDDOC  \cite{li2024prompt}      & 40.9          & 35.0          & 59.0          & 21.3          & 40.4          & 29.9          & 42.9          & 38.5          \\
UFR   \cite{liu2024unbiased}       & 43.6          & 38.1          & 66.1          & 14.7          & 49.1          & 26.4          & 42.9          & 36.9          \\ \hline
Ours         & \textbf{47.8} & \underline{40.1}          & \textbf{71.8} & \textbf{27.2} & \textbf{55.2} & \textbf{38.8} & \textbf{50.1} & \textbf{47.3} \\ \hline
\end{tabular}
\end{table}

\begin{table}[]
\centering
\setlength\tabcolsep{2.5pt}
\footnotesize
\caption{Quantitative results (\%) on the Dusk-Rainy scene.
} \label{tab:dusk_rainy}
\begin{tabular}{l|ccccccc|
>{\columncolor[HTML]{d1d1d1}}c }
\hline
Method       & bus           & bike          & car           & motor         & person        & rider         & truck         & mAP           \\ \hline
Faster R-CNN \cite{ren2016faster} & 34.6          & 22.3          & 63.6          & 7.9           & 24.2          & 13.2          & 39.4          & 29.3          \\
IBN-Net \cite{pan2018two}     & 37.0          & 14.8          & 50.3          & 11.4          & 17.9          & 13.3          & 38.4          & 26.1          \\
SW    \cite{pan2019switchable}       & 35.2          & 16.7          & 50.1          & 10.4          & 20.1          & 13.0          & 38.8          & 26.3          \\
IterNorm \cite{huang2019iterative}     & 32.9          & 14.1          & 38.9          & 11.0          & 15.5          & 11.6          & 35.7          & 22.8          \\
ISW   \cite{choi2021robustnet}       & 34.7          & 16.0          & 50.0          & 11.1          & 17.8          & 12.6          & 38.8          & 25.9          \\
S-DGOD  \cite{wu2022single}     & 37.1          & 19.6          & 50.9          & 13.4          & 19.7          & 16.3          & 40.7          & 28.2          \\
CLIP-Gap \cite{vidit2023clip}    & 37.8          & 22.8          & 60.7          & 16.8          & 26.8          & 18.7          & 42.4          & 32.3          \\
G-NAS   \cite{wu2024g}     & \underline{44.6}    & 22.3          & 66.4          & 14.7          & \underline{32.1}    & \underline{19.6}    & \underline{45.8}    & \underline{35.1}    \\
PDDOC   \cite{li2024prompt}     & 39.4          & \underline{25.2}    & 60.9          & \textbf{20.4} & 29.9          & 16.5          & 43.9          & 33.7          \\
UFR   \cite{liu2024unbiased}       & 37.1          & 21.8          & \underline{67.9}    & 16.4          & 27.4          & 17.9          & 43.9          & 33.2          \\ \hline
Ours         & \textbf{50.3} & \textbf{30.9} & \textbf{75.7} & \underline{19.6}    & \textbf{40.4} & \textbf{26.4} & \textbf{52.9} & \textbf{42.3} \\ \hline
\end{tabular}
\end{table}

\begin{table}[]
\centering
\setlength\tabcolsep{2.5pt}
\footnotesize
\caption{Quantitative results (\%) on the Night-Rainy scene.
} \label{tab:night_rainy}
\begin{tabular}{l|ccccccc|
>{\columncolor[HTML]{d1d1d1}}c }
\hline
Method       & bus           & bike          & car           & motor         & person        & rider         & truck         & mAP           \\ \hline
Faster R-CNN \cite{ren2016faster} & 19.8          & 8.6           & 30.2          & 0.3           & 0.9           & 4.8           & 16.6          & 12.8          \\
IBN-Net \cite{pan2018two}     & 24.6          & 10.0          & 28.4          & 0.9           & 8.3           & 9.8           & 18.1          & 14.3          \\
SW   \cite{pan2019switchable}        & 22.3          & 7.8           & 27.6          & 0.2           & 10.3          & 10.0          & 17.7          & 13.7          \\
IterNorm  \cite{huang2019iterative}   & 21.4          & 6.7           & 22.0          & 0.9           & 9.1           & 10.6          & 17.6          & 12.6          \\
ISW   \cite{choi2021robustnet}       & 22.5          & 11.4          & 26.9          & 0.4           & 9.9           & 9.8           & 17.5          & 14.1          \\
S-DGOD   \cite{wu2022single}    & 24.4          & 11.6          & 29.5          & \underline{9.8}           & 10.5          & 11.4          & 19.2          & 16.6          \\
CLIP-Gap  \cite{vidit2023clip}   & 28.6          & \underline{12.1}    & 36.1          & 9.2           & 12.3          & 9.6           & 22.9          & 18.7          \\
G-NAS   \cite{wu2024g}     & 28.6          & 9.8           & \underline{38.4}    & 0.1           & 13.8          & 9.8           & 21.4          & 17.4          \\
PDDOC  \cite{li2024prompt}      & 25.6          & \underline{12.1}    & 35.8          & \textbf{10.1} & \underline{14.2}    & \underline{12.9}    & 22.9          & \underline{19.2}    \\
UFR   \cite{liu2024unbiased}       & \underline{29.9}    & 11.8          & 36.1          & 9.4           & 13.1          & 10.5          & \underline{23.3}    & \underline{19.2}    \\ \hline
Ours         & \textbf{34.3} & \textbf{14.2} & \textbf{49.6} & 2.1           & \textbf{17.2} & \textbf{15.6} & \textbf{30.6} & \textbf{23.4} \\ \hline
\end{tabular}
\end{table}

\begin{table}[]
\centering
\setlength\tabcolsep{2.5pt}
\footnotesize
\caption{Quantitative results (\%) on the Daytime-Foggy scene.
} \label{tab:daytime-foggy}
\begin{tabular}{l|ccccccc|
>{\columncolor[HTML]{d1d1d1}}c }
\hline
Method       & bus           & bike          & car           & motor         & person        & rider         & truck         & mAP           \\ \hline
Faster R-CNN \cite{ren2016faster} & 30.2          & 26.6          & 54.9          & 27.2          & 35.1          & 37.6          & 19.4          & 33.0          \\
IBN-Net  \cite{pan2018two}    & 29.9          & 26.1          & 44.5          & 24.4          & 26.2          & 33.5          & 22.4          & 29.6          \\
SW   \cite{pan2019switchable}        & 30.6          & 36.2          & 44.6          & 25.1          & 30.7          & 34.6          & 23.6          & 30.8          \\
IterNorm  \cite{huang2019iterative}   & 29.7          & 21.8          & 42.4          & 24.4          & 26.0          & 33.3          & 21.6          & 28.4          \\
ISW   \cite{choi2021robustnet}       & 29.5          & 26.4          & 49.2          & 27.9          & 30.7          & 34.8          & 24.0          & 31.8          \\
S-DGOD  \cite{wu2022single}     & 32.9          & 28.0          & 48.8          & 29.8          & 32.5          & 38.2          & 24.1          & 33.5          \\
CLIP-Gap  \cite{vidit2023clip}   & 36.2          & 34.2          & 57.9          & \underline{34.0}    & 38.7          & \underline{43.8}          & 25.1          & 38.5          \\
G-NAS   \cite{wu2024g}     & 32.4          & 31.2          & 57.7          & 31.9          & 38.6          & 38.5          & 24.5          & 36.4          \\
PDDOC   \cite{li2024prompt}     & 36.1          & \underline{34.5}    & 58.4          & 33.3          & \underline{40.5}    & \textbf{44.2} & \underline{26.2}          & 39.1          \\
UFR   \cite{liu2024unbiased}       & \textbf{36.9} & \textbf{35.8} & \underline{61.7}    & 33.7          & 39.5          & 42.2          & \textbf{27.5} & \underline{39.6}    \\ \hline
Ours         & \underline{36.6}    & 33.2          & \textbf{62.6} & \textbf{34.6} & \textbf{42.9} & \textbf{44.2} & \textbf{27.5} & \textbf{40.2} \\ \hline
\end{tabular}
\end{table}

\textbf{Results on Night-Clear Scene.} Table \ref{tab:night_clear} summarizes the results for the Night-Clear scene. Due to reduced visibility at night, object detection becomes significantly more challenging compared with Daytime-Clear. In this scenario, the proposed method achieves the best performance with 47.3\% mAP, outperforming previous SOTA methods across six object categories, except for \textit{bike}. Low light conditions in night scenes lead to confusion between visually similar categories, such as \textit{bike} and \textit{motor}.

\textbf{Results on Dusk-Rainy Scene.} Table \ref{tab:dusk_rainy} summarizes the results for the Dusk-Rainy scene, where low light and rain affect detection performance. The proposed method outperforms others, showing significant improvements across most categories, especially \textit{bike}, \textit{person}, and \textit{rider}, which are prone to occlusion from small objects like those influenced by rain or fog. However, confusion between \textit{motor} and \textit{bike} slightly lowers \textit{motor} performance.

\textbf{Results on Night-Rainy Scene.} Table \ref{tab:night_rainy} summarizes the results for the Night-Rainy scene. The Night-Rainy scene is the most challenging condition owing to low light and the effects of rain. Despite these challenges, our method outperforms others, achieving the highest performance. However, \textit{motor} exhibits lower performance due to confusion with \textit{bike}.

\textbf{Results on Daytime-Foggy Scene.} Table \ref{tab:daytime-foggy} summarizes the results for the Daytime-Foggy scene, where fog causes occlusions and blurring, complicating detection. The proposed method outperforms others across most categories. However, foggy conditions increase confusion between visually similar categories like \textit{bus} and \textit{truck}, and \textit{bike} and \textit{motor}.

\begin{table}[htbp]
\centering
\setlength\tabcolsep{2.1pt}
\footnotesize
\caption{Ablation study results (\%) of the proposed CD-FKD. DC, NC, DR, NR, and DF represent Daytime-Clear, Night-Clear, Dusk-Rainy, Night-Rainy, and Daytime-Foggy, respectively.}
\label{tab:ablation}
\begin{tabular}{l|cc|ccccc|c}
\hline
Methods &  Corrupt\&Down & FKD & DC            & NC            & DR            & NR            & DF            & Avg.          \\ \hline
Faster R-CNN      &      {\color{red}\faTimes}            &    {\color{red}\faTimes}          & 54.9          & 36.6          & 27.9          & 12.1          & 32.1          & 27.2          \\ \hline
Ours    &        {\color{blue}\faCheck}           &    {\color{red}\faTimes}          & 58.8          & 41.1          & 33.9          & 14.9          & 34.0          & 30.9          \\
Ours    &       {\color{blue}\faCheck}           &    $\mathcal{L}_{\text{glo}}$          & 62.0          & 46.3          & 41.7          & 21.4          & 39.5          & 37.2          \\
Ours    &        {\color{blue}\faCheck}          &    $\mathcal{L}_{\text{ins}}$          & 62.5          & 46.7          & 42.1          & 21.9          & 39.7          & 37.6          \\
Ours    &          {\color{blue}\faCheck}        &    $\mathcal{L}_{\text{glo}}$+$\mathcal{L}_{\text{ins}}$      & \textbf{62.7} & \textbf{47.3} & \textbf{42.3} & \textbf{23.4} & \textbf{40.2} & \textbf{38.3} \\ \hline
\end{tabular}
\hspace{0.05\textwidth}
\centering
\setlength\tabcolsep{4.3pt}
\footnotesize
\caption{Ablation study on the source domain test set compares Faster R-CNN and CD-FKD, evaluating CD-FKD with only corruption (Corrupt) and with both corruption and downscaling (Corrupt\&Down).}
\label{tab:scale}
\begin{tabular}{l|ccccc}
\hline
Method       & mAP  & mAP$_{\textit{50}}$ & mAP$_{\textit{s}}$ & mAP$_{\textit{m}}$ & mAP$_{\textit{l}}$ \\ \hline
Faster R-CNN   & 29.5 & 54.9    & 8.2   & 32.4   & 54.6   \\
+ Corrupt    & 34.5 & 62.1    & 13.2   & 37.7   & \textbf{57.6}   \\
+ Corrupt\&Down & \textbf{34.8} & \textbf{62.7}    & \textbf{13.6}   & \textbf{38.0}   & \textbf{57.6}   \\ \hline
\end{tabular}
\end{table}

\subsection{Ablation Study}
We conducted an ablation study to analyze the impact of different components of the proposed method. First, we evaluated the effects of corruption and downscaling (Corrupt\&Down) on the input data. Second, we performed an ablation analysis to evaluate the impact of the FKD, specifically $\mathcal{L}_{\textit{global}}$ and $\mathcal{L}_{\textit{instance}}$.

Table \ref{tab:ablation} summarizes the results. "Avg." represents the average performance across the four target domains. The application of corruption and downscaling to the source images significantly improved performance compared to the baseline, demonstrating its effectiveness in enhancing generalization. Moreover, using both $\mathcal{L}_{\textit{global}}$ and $\mathcal{L}_{\textit{instance}}$ distillation methods resulted in a 38.3\% mAP, a notable improvement over the baseline. Each distillation method individually contributed to the performance gains, not only for the target domains but also for the source domain. Through distillation, our model learned to replicate the feature representations from the teacher network, improving adaptability to the source domain while avoiding overfitting to the diversified data.

Table \ref{tab:scale} presents the results of another ablation study, examining the effect of corruption and downscaling on the input images. The proposed method, when applied to the student network, significantly outperformed the baseline. Notably, the downscaling technique achieved a 0.4\% improvement in mAP$_{\textit{s}}$ and a 0.3\% improvement in mAP$_{\textit{m}}$ compared to corruption-only experiments, underscoring the method’s effectiveness in improving small object detection.

\begin{figure}[t]
\centering
  \includegraphics[width=0.90\columnwidth]{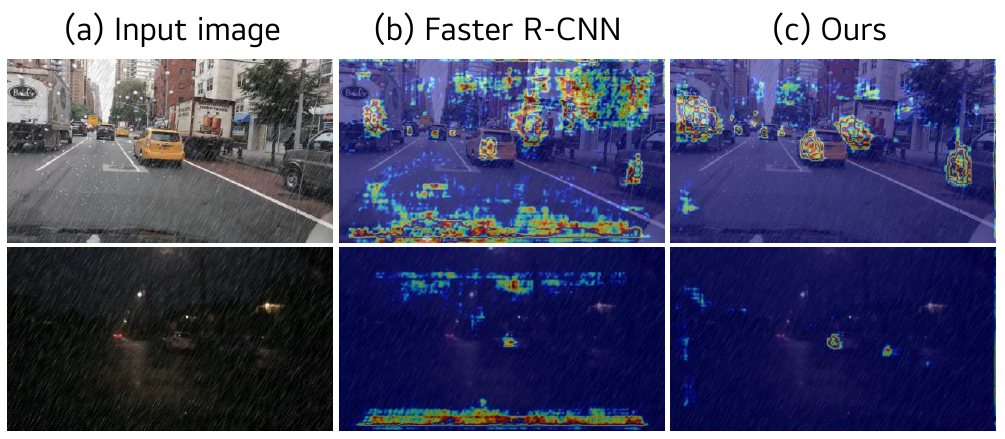}
  \caption{Heatmap visualization for target domain scenes. The left column displays the original images, the middle column presents the results from Faster R-CNN, and the right column shows the results from our method. The heatmaps highlight the areas where the model focuses, with regions of higher attention marked by a redder hue.} 
  \label{fig:feat_vis}
\end{figure}

Additionally, Figure \ref{fig:feat_vis} compares the heatmaps generated by the baseline and the proposed CD-FKD. The baseline-generated heatmap mainly highlights irrelevant background areas. In contrast, the proposed method reduces background focus and places more attention on the objects. This shows that the proposed approach focuses on objects instead of the background in unseen domains, thereby improving generalization performance.

\begin{table}[htbp]
\centering
\footnotesize
\caption{The impact of the hyperparameters $\alpha$ and $\beta$ of $\mathcal{L}_{\textit{global}}$ and $\mathcal{L}_{\textit{instance}}$ in the distillation training phase.} \label{tab:loss balance}
\begin{tabular}{l|llllll}
\hline
$\alpha$/$\beta$ & DC   & NC   & DR   & NR   & DF   & Avg. \\ \hline
0.5/1.5    & 62.5 & 46.8 & 42.1 & 23.1 & 40.1 & 38.0 \\
\textbf{1.0/1.0}     & \textbf{62.7} & \textbf{47.3} & \textbf{42.3} & \textbf{23.4} & \textbf{40.2} & \textbf{38.3} \\
1.5/0.5     & 62.1 & 46.7 & 41.8 & 22.9 & 40.1 & 37.9 \\ \hline
\end{tabular}
\end{table}

In Table \ref{tab:loss balance}, we applied the hyperparameters $\alpha$ and $\beta$ to the $\mathcal{L}_{\textit{global}}$ and $\mathcal{L}_{\textit{instance}}$ loss functions during the feature distillation training phase from the teacher network to the student network. According to the results from the ablation study in the main manuscript, when only one of the two loss functions ($\mathcal{L}_{\textit{global}}$ or $\mathcal{L}_{\textit{instance}}$) was used, $\mathcal{L}_{\textit{instance}}$ contributed slightly more to performance improvement. However, the final results show that applying both $\mathcal{L}_{\textit{global}}$ and $\mathcal{L}_{\textit{instance}}$ resulted in the largest performance gain.

Therefore, we further analyze the effects of both loss functions. Table \ref{tab:loss balance} reports the performance achieved by applying $\alpha$ and $\beta$ values of 0.5/1.5, 1.0/1.0, and 1.5/0.5. Our evaluation shows that the optimal $\alpha$ and $\beta$ values of 1.0 and 1.0 strike an effective balance between transferring global features through $\mathcal{L}_{\textit{global}}$ and instance-wise features through $\mathcal{L}_{\textit{instance}}$. This balance maximized the learning efficiency of the student network.

\section{Conclusion}

In this study, we introduced a novel approach for single-domain generalized object detection using CD-FKD. Our proposed method effectively addressed the domain shift problem and significantly improved generalization to unseen target domains using only a single source domain. We applied global and instance-wise feature distillation, enabling the model to extract robust and meaningful object-centric features while maintaining detection performance even under severe corruptions. Experimental results demonstrated that CD-FKD consistently outperformed existing methods under adverse weather conditions. Furthermore, an ablation study confirmed the effectiveness of each component in the proposed method, leading to performance improvements in both unseen target and source domains.

\section*{Acknowledgement}
\scriptsize{
This research was supported by the National Research Council of Science \& Technology (NST) grant by the Korea government (MSIT) (No. GTL25041-000)
}
% \end{spacing}

\begin{spacing}{0.9} % 'setspace' 패키지 필요, 줄 간격을 0.9배로 압축
\footnotesize
\bibliography{references.bib}{}
\bibliographystyle{IEEEtran}
\end{spacing}

\end{document}